%% file: main.tex
\documentclass[10pt,twocolumn,letterpaper]{article}

\usepackage[pagenumbers]{cvpr} 
\input{preamble}
\definecolor{cvprblue}{rgb}{0.21,0.49,0.74}

\usepackage[pagebackref,breaklinks,colorlinks,allcolors=cvprblue]{hyperref}
\usepackage{graphicx}

\def\paperID{arXiv}
\def\confName{CVPR}
\def\confYear{2026}

\title{Revisiting Salient Object Detection from an Observer-Centric Perspective}

\author{
Fuxi Zhang,
Yifan Wang,
Hengrun Zhao,
Zhuohan Sun,
Changxing Xia,
Lijun Wang,\\
Huchuan Lu,
Yangrui Shao,
Chen Yang,
Long Teng
\\[0.6em]
Dalian University of Technology
}

\begin{document}


\maketitle

\input{sec_arxiv/0_abstract}    
\input{sec_arxiv/1_intro}
\input{sec_arxiv/2_related}

\input{sec_arxiv/3_dataset}

\input{sec_arxiv/4_method}
\input{sec_arxiv/5_experiment}
\input{sec_arxiv/6_conclusion}

\clearpage
{
    \small
    \bibliographystyle{ieeenat_fullname}
    \bibliography{main}
}

\input{sec_arxiv/X_suppl}

\end{document}

%% file: preamble.tex
\usepackage{multirow}
\usepackage{algorithm}
\usepackage{algorithmic}


%% file: sec_arxiv/0_abstract.tex
\begin{abstract}
Salient object detection is inherently a subjective problem, as observers with different priors may perceive different objects as salient. However, existing methods predominantly formulate it as an objective prediction task with a single groundtruth segmentation map for each image, which renders the problem under-determined and fundamentally ill-posed.
To address this issue, we propose \textbf{O}bserver-\textbf{C}entric \textbf{S}alient \textbf{O}bject \textbf{D}etection (\textbf{OC-SOD}), where salient regions are predicted by considering not only the visual cues but also the observer-specific factors such as their preferences or intents. 
As a result, this formulation captures the intrinsic ambiguity and diversity of human perception, enabling personalized and context-aware saliency prediction.
By leveraging multi-modal large language models, we develop an efficient data annotation pipeline and construct the first OC-SOD dataset named OC-SODBench, comprising 33k training, validation and test images with 152k textual prompts and object pairs. Built upon this new dataset, we further design OC-SODAgent, an agentic baseline which performs OC-SOD via a human-like ``Perceive-Reflect-Adjust" process. Extensive experiments on our proposed OC-SODBench have justified the effectiveness of our contribution. Through this observer-centric perspective, we aim to bridge the gap between human perception and computational modeling, offering a more realistic and flexible understanding of what makes an object truly “salient.” Code and dataset are publicly available at: \url{https://github.com/Dustzx/OC_SOD}
\end{abstract}

%% file: sec_arxiv/1_intro.tex
\section{Introduction}
\label{sec:intro}
Salient Object Detection (SOD) aims to model the human visual attention mechanism by identifying and segmenting the most visually prominent objects in a scene. As a long-standing and fundamental problem in computer vision, it has attracted extensive interest from the community and found wide applications in real-world scenarios, such as image enhancement~\cite{image-enhance}, image retrieval~\cite{sodImageRetr}, autonomous driving~\cite{sodAutoDrive}, and robotic perception~\cite{sodRobotperception}, among others. 

\begin{figure}[t]
  \centering  \includegraphics[width=1.0\linewidth]{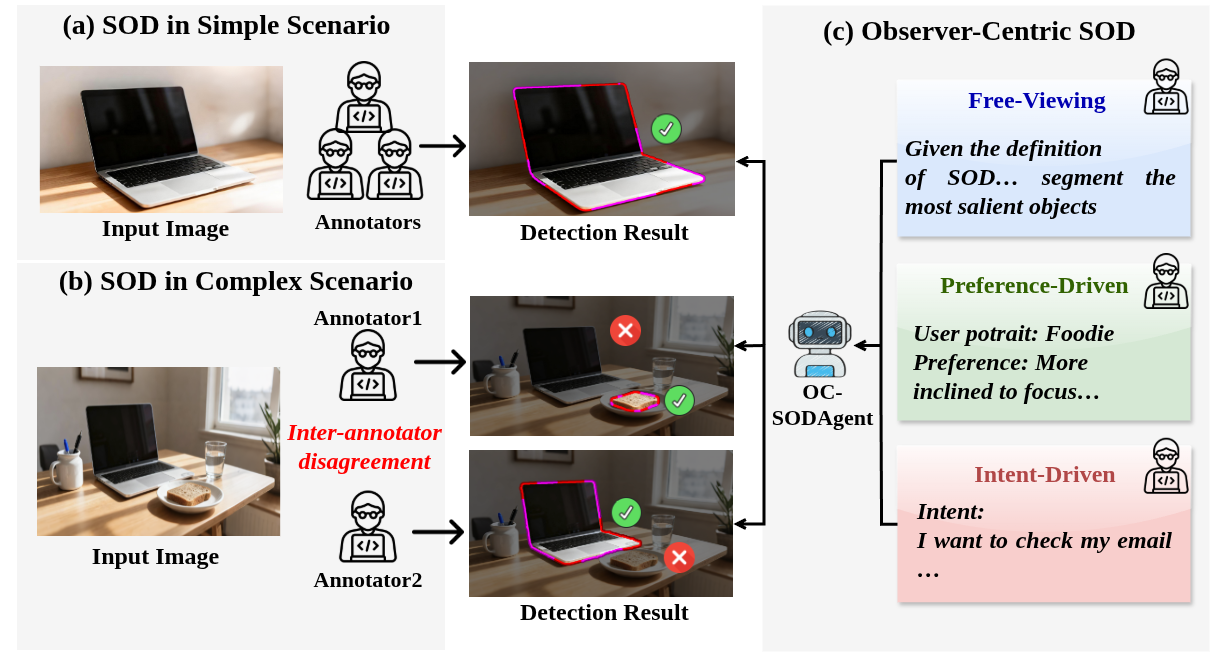}
  \caption{Illustration of traditional Salient Object Detection (SOD) limitations and our proposed Observer-Centric (OC-SOD) solution. (a) Traditional SOD aligns with human consensus in simple scenes. (b) In complex scenes, it becomes ill-posed due to inter-annotator disagreements driven by diverse subjective priors. (c) Our OC-SOD paradigm resolves this ambiguity by modeling distinct subjective contexts. This includes the ``Free-viewing mode"—a feature-driven mode, as well as modes defined by specific subjective priors, such as Preference-Driven (e.g., ``Foodie") and Intent-Driven (e.g., ``I want to check my email"). Integrating these explicit priors renders the segmentation task well-posed and unambiguous.}
  \label{fig:fig1}
\end{figure}



Though significant progress has been achieved in the past decade, a fundamental contradiction still remains between the ambiguous definition of visual saliency and the prevailing formulation of SOD. On the one hand, visual saliency is inherently a subjective concept and largely shaped by observer-specific factors, \ie, different observers driven by distinct interests and purposes often hold varied definitions of saliency even for the same input image (\cf \cref{fig:fig1} (b)). On the other hand, existing methods mostly pose SOD as an objective and deterministic prediction problem, where salient regions are inferred purely via visual cues and a single, definite groundtruth output is assumed for each input image. While such a formulation may suffice for simple scenes dominated by a single salient object (\eg, \cref{fig:fig1} (a)), it often fails in complex or ambiguous settings. 
thereby leading to the well-known ill-posed issue of existing SOD research~\cite{salient-inconsistency,psod,vst,vscode}.

As illustrated in \cref{fig:fig1}(b), when an image contains both a piece of bread and a laptop, human annotators may even fail to reach a consensus on which object is more salient. The bread may appear more salient to a food enthusiast (reflecting long-term preference) or to a hungry individual (reflecting immediate intent), whereas the laptop may attract a tech enthusiast (long-term preference) or someone about to send an email (immediate intent). These observer-dependent and context-driven variations are largely beyond the modeling capacity of existing SOD methods.

Motivated by the above observations, we propose OC-SOD, a comprehensive paradigm for studying SOD from an Observer-Centric perspective.
As shown in \cref{fig:fig1}(c), OC-SOD investigates three representative modes: free-viewing, preference-driven, and intent-driven. The free-viewing mode simulates an observer without prior preference, identifying salient regions based on visual cues, which aligns with existing SOD approaches and remains effective in simple scenarios. In contrast, the preference- and intent-driven modes aim to detect salient regions conditioned on a description of the observer’s cognitive state, which encapsulates their long-term interests and short-term purposes, respectively, within a given context. By reformulating SOD from this observer-centric perspective, OC-SOD provides a principled way to model the subjective nature of visual saliency, thereby making saliency prediction less ambiguous and more personalized to individual observers, and alleviating the ill-posed nature of conventional SOD paradigms.

Existing SOD datasets lack subjective, observer-specific annotations, and thus cannot be directly used to facilitate the study of OC-SOD. To address this limitation, we develop an efficient data annotation pipeline powered by Multi-modal Large Language Models (MLLMs). Through scene analysis, prompt annotation, data verification, and manual curation, we construct OC-SODBench dataset comprising 33k images and 152k instruction-mask pairs. Each instruction provides a textual description of the OC-SOD task along with the observer’s current cognitive state, which corresponds to a specific salient region in the input image. To the best of our knowledge, this is the first dataset that can enables both training and evaluating OC-SOD models. 

Building upon this dataset, we further propose OC-SODAgent, an agentic baseline model. At its core, OC-SODAgent employs an MLLM to interpret the input instruction, reason about saliency from the observer’s perspective, and invoke a pre-trained SAMv2 model~\cite{sam2} to segment the salient regions. The final saliency prediction is obtained through an iterative refinement process that mimics a human-like ``Perceive–Reflect–Adjust" manner. Without fine-tuning, OC-SODAgent already surpasses existing MLLMs by a substantial margin. Moreover, fine-tuning on the OC-SODBench training set yields consistent performance improvements across all compared methods, further validating the effectiveness and general utility of our proposed dataset.



The contribution of this paper can be summarized into three-fold:
\begin{itemize}
\item \textbf{Observer-centric reformulation of SOD:} We introduce OC-SOD, a novel paradigm that reformulates SOD from an observer-centric perspective, explicitly modeling the subjective nature of visual saliency conditioned on individual preferences and intents.

\item \textbf{Efficient annotation pipeline and new benchmark:} We develop an MLLM-assisted annotation pipeline and construct OC-SODBench, a dataset comprising 33K images and 152K instruction–mask pairs, enabling both training and evaluation of observer-centric SOD models.

\item \textbf{Agentic baseline model:} We propose OC-SODAgent, an agentic MLLM-based framework that reasons about saliency from the observer’s perspective and iteratively refines predictions, which surpasses existing MLLMs by a considerable margin on OC-SODBench.
\end{itemize}

We hope our work can bridge the divide between human perception and computational modeling of saliency, paving the way toward more human-aligned and context-aware understanding of visual saliency. Source code and datasets will be released upon publication.

%% file: sec_arxiv/2_related.tex
\section{Related Work}
\label{sec:related}
\subsection{Salient Object Detection Benchmark}

The field of Salient Object Detection (SOD) has evolved significantly through its benchmarks. Initial studies, relying on datasets like MSRA~\cite{msra}, defined saliency as an objective visual property, using low-level cues to identify the most prominent object. This is followed by benchmarks such as ECSSD~\cite{ecssd}, HKU-IS~\cite{hku-is}, PASCAL-S~\cite{pascal-s}, and DUT-OMRON~\cite{dut-omron}, which presented more challenging, cluttered, and realistic scenes. The DUTS dataset~\cite{duts} further scales the data for training deep learning models.
Despite the advances, these SOD datasets typically provide a single ground-truth mask per image. implicitly assuming a universal standard of saliency. This design neglects the inherent subjectivity of human attention, where different observers may focus on different objects based on their context, preferences, or interests.

To alleviate the above issue, several task variants have been proposed~\cite{sos,isod,sor,psod}. Among them, Salient Object Ranking (SOR)~\cite{sor,mllmsor} learns relative saliency between objects. Pluralistic SOD (PSOD)~\cite{psod} is proposed to capture saliency diversity by generating multiple plausible masks and preference scores. 
Despite this advancement, these approaches essentially overlook the subjective diversity of human perception, and are unable to infer the salient objects of potential interest in a scene from the observer's perspective based on their interests or intents.

\subsection{Reasoning Segmentation}

Recent years have witnessed the advent of Multimodal Large Language Models (MLLMs)~\cite{llava}, which excel at aligning textual semantics with visual information. Therefore, it becomes possible to integrate human intent and subjective reasoning into vision tasks.
Building on this idea, reasoning segmentation establishes a new paradigm for language-driven segmentation. LISA~\cite{lisa} pioneers this direction by coupling an MLLM with SAM~\cite{sam} for segmentation from implicit textual cues.
LLM-Seg~\cite{llmseg} extends this framework with modular components (including LLaVA \cite{llava}, DINOv2~\cite{oquab2023dinov2}, and SAM \cite{sam}) and releases a large dataset termed LLM-Seg40k. 
The scope is further advanced by PixelLM~\cite{pixellm}, which handles multi-object scenarios, while MMR~\cite{mmr} refines the paradigm to achieve part-level understanding with PACO-LVIS~\cite{paco}.


While these works move toward semantic reasoning and contextual understanding, they primarily operate under the paradigm of explicit visual-text alignment, which inherently limits their capacity to model the subjective and dynamic nature of human attention. Consequently, these approaches remain limited in addressing the observer-centric aspects of saliency perception.

%% file: sec_arxiv/3_dataset.tex
\begin{figure*}[t]
  \centering
    \includegraphics[width=0.75\linewidth]{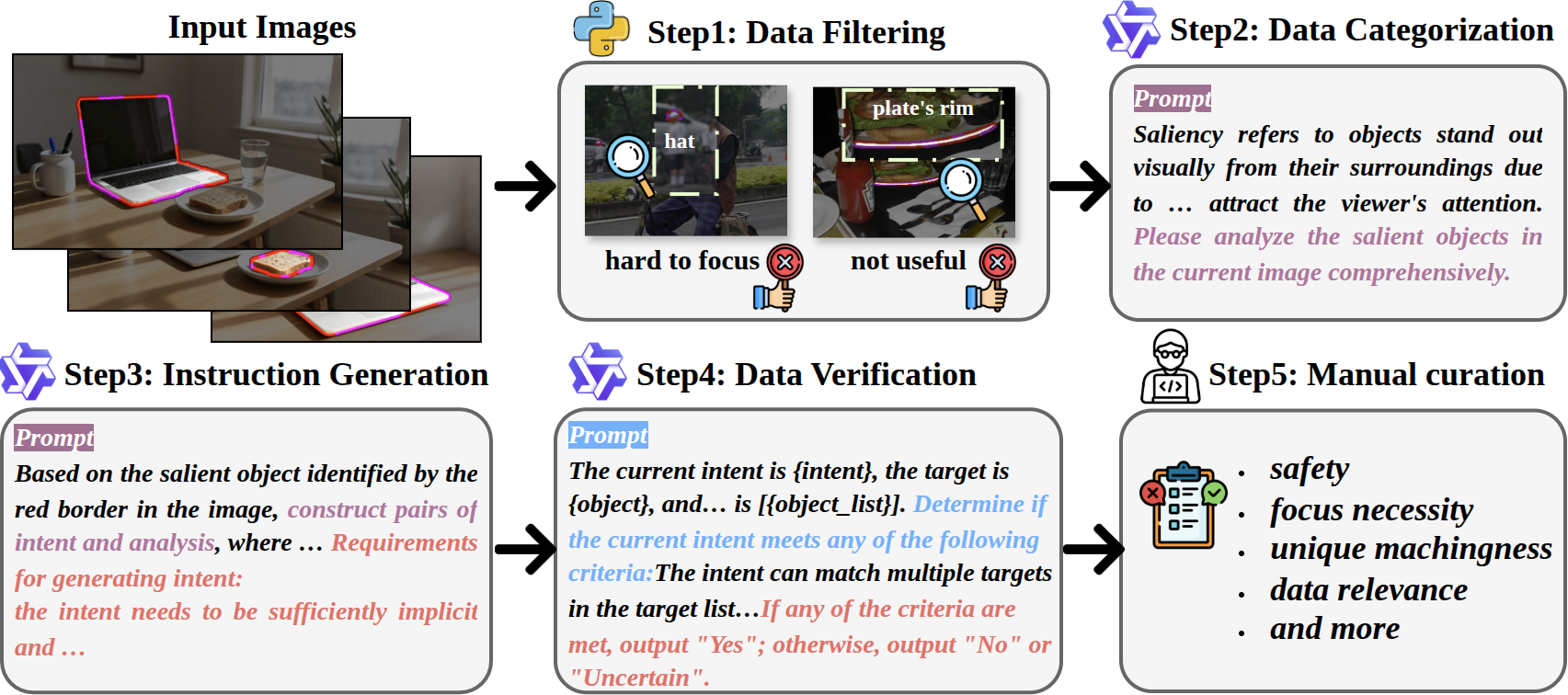}
    \caption{An overview of our 5-step data annotation pipeline. The process begins with pre-annotated images and employs Multimodal Large Language Models (MLLMs) for key tasks. The steps are: (1) Data Filtering to remove unsuitable samples (e.g., ``hard to focus," ``not useful"); (2) MLLM-driven Data Categorization to analyze saliency; (3) Instruction Generation using an MLLM to create intent- or preference-based prompts; (4) automated Data Verification by an MLLM to check for errors; and (5) final Manual Curation by experts to ensure dataset quality, checking criteria such as safety, focus necessity, and relevance.}
  \label{fig:pipeline}
\end{figure*}
\section{Observer-Centric Salient Object Detection}
\label{sec:dataset}

\subsection{The OC-SOD Task Setting}
\label{sec:task_setting}
Traditional Salient Object Detection (SOD) treats saliency as an objective property of an image, assuming a single ground truth map shared across all observers: $M = P(I)$, where the detection model $P(\cdot)$ only takes the image as input and predict the saliency map $M$. This formulation, however, overlooks the inherently subjective and context-dependent nature of human visual attention. By collapsing perceptual diversity into a single deterministic target, they render the problem ill-posed and perceptually inconsistent with human cognition.

To overcome this limitation, we propose Observer-Centric Salient Object Detection (OC-SOD), which reformulates saliency prediction as a conditional generation problem guided by observer-specific factors: 
\begin{equation}
    M, D = P(I|T),
\end{equation}
where $T$ represents the textual instructions describing the task setting and observer's subjective states, and $D$ optionally captures the model’s intermediate reasoning or explanatory process. By explicitly modeling the observer-related cues, saliency prediction task becomes less ambiguous and more personalized to individual observers. In this work, we instantiate OC-SOD through three representative observation modes: free-viewing, preference-driven, and intent-driven, to concretize the paradigm. Nevertheless, other observer-centric conditions can be readily incorporated within our flexible formulation. Please refer to the supplementary materials for detailed instruction templates of all the observation modes.

\textbf{Free-Viewing Mode.} This mode represents observers without explicit preferences or intentions. It aligns with the conventional definition of SOD, where saliency emerges purely from intrinsic visual cues such as contrast, color, and spatial distinctiveness (\ie, $T=\emptyset$). Although limited in capturing subjectivity, it remains effective for scenes dominated by a single visually prominent object.

\textbf{Preference-Driven Mode.} This mode reflects an observer’s enduring interests or habitual tendencies, which shape their visual attention even in neutral viewing contexts. For instance, a person who identifies as a “food lover” would naturally focus on edible items in a scene, whereas a “technology enthusiast” would be drawn to electronic devices. By modeling such stable inclinations, this mode accounts for the inter-observer variability that underlies perceptual differences across individuals.

\textbf{Intent-Driven Mode.} This mode captures momentary goals or situational purposes that dynamically modulate attention. When viewing the same scene, an observer who intends to “check email” may prioritize the laptop as salient, while another whose intent is to “prepare breakfast” may instead attend to the bread. This conditional formulation allows OC-SOD to flexibly adapt saliency to different contextual demands.

\subsection{The OC-SODBench Dataset}
\label{sec:dataset_intro}
Since there is no existing datasets that can fulfill the needs of training and testing OC-SOD methods, we develop an efficient data annotation pipeline powered by MLLMs. With limited manual laboring, we construct the OC-SODBench dataset using this annotation pipeline. 

\subsubsection{Data Annotation Pipeline}
While existing Multi-modal Large Language Models (MLLMs) demonstrate strong capabilities in image understanding and reasoning, they remain limited in generating precise pixel-level segmentation results. Therefore, to reduce manual laboring, our data annotation pipeline is built upon existing saliency detection and image segmentation datasets that already contain pixel-level ground truth masks (We name these datasets as pre-annotated). And the MLLMs are primarily employed for data categorization, instruction generation, and quality verification. The overall pipeline is organized into the following 5 key steps.    

\textbf{Data Filtering.} Since not all the aforementioned pre-annotated datasets are designed for saliency detection purposes, we first perform a rule-based data filtering step to remove unsuitable samples. the target objects that meet one of the following two criteria will be deemed as unsuitable for OC-SOD and excluded: 1) Difficult to focus. Objects that occupy a very small area (less than 0.1\% of the image) or appear within dense clusters of visually similar instances, where distinguishing a single salient target becomes ambiguous. 2) Semantically uninformative. This refers to regions lacking clear semantic meaning or practical functionality, such as part-level concepts (e.g., “plate’s rim”) or background fragments, which provide little perceptual relevance. The above criteria can be efficiently verified using the available ground-truth category labels and segmentation masks, ensuring that only semantically valid and visually discernible objects are retained.

\textbf{Data Categorization.} For each image, we employ the Qwen3-VL-Instruct model to perform detailed captioning, describing both the global scene context and the prominent foreground objects. Based on these captions, the MLLM is instructed to determine whether the image contains a single, unambiguous salient object and provide reasoning to justify its judgment. Images meeting this condition are categorized as free-viewing mode, as their saliency can be inferred purely from visual cues.
Images containing multiple plausible salient targets or exhibiting more complex scenarios are instead routed to either the preference-driven or intent-driven modes, which require subjective interpretation. Among these complex cases, samples where the target corresponds to object parts are explicitly assigned to the intent-driven mode, since intent more naturally aligns with functional or part-level focus. The remaining samples are randomly assigned to either the preference-driven or intent-driven categories.

\textbf{Instruction Generation.} Images belonging to the free-viewing mode will be annotated with a fixed instruction ``Identify and segment the most salient regions according to visual context, color contrast, semantic meaning." For objects belonging to preference-driven mode, we provide the MLLM with the input image along with the object segmentation mask, and prompt it to generate a textual description of interesting observers' portrait, including their specific interest, preference, \etc. The final instruction is formulated as ``Here is the observer's portrait $<portrait>$. Identify and segment the most salient regions according to the observer's interest and preference.", where $<portrait>$ denotes the generated description. Similarly, for objects belonging to intent-driven mode, the MLLM is guided to generate an observer’s immediate intent for focusing on a particular target. The resulting instruction takes the form: ``Here is the observer’s intent $<intent>$. Identify and segment the most salient regions according to this intent." All prompt templates used for data categorization and instruction generation are included in the Supplementary Materials.
\begin{figure*}[t]
  \centering
    \includegraphics[width=1.0\linewidth]{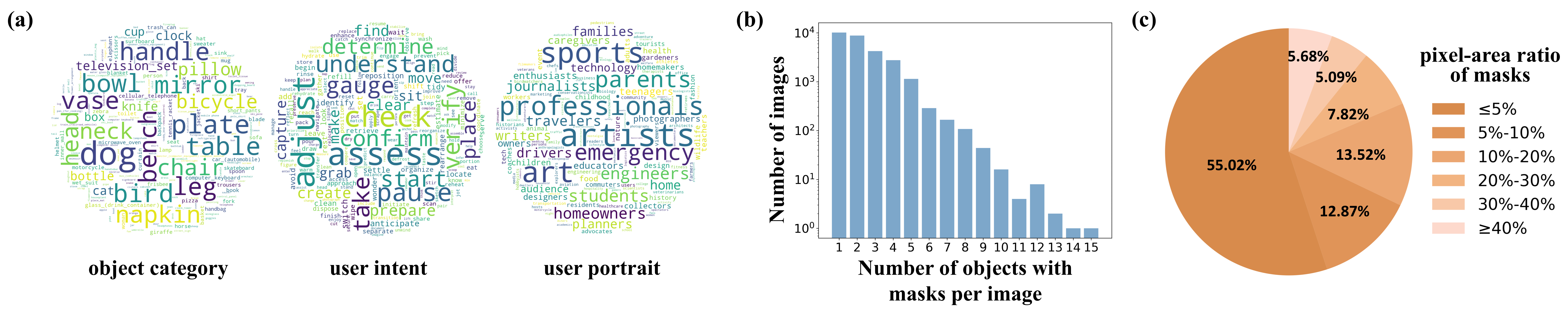}
    \vspace{-5mm}
        \caption{Statistical overview of the OC-SODBench dataset. (a) Word clouds of target objects, preferences, and intents. (b) Histogram of object counts. (c) The pixel-area ratio of target masks. Please zoom in for details.}
  \label{fig:statistics}
\end{figure*}

\textbf{Data Verification and Manual Curation.}
To further improve the data quality, we introduce an automated verification step followed by manual curation. Specifically, we employ the Qwen3-VL-Thinking model and provide it with the image, the corresponding annotations (both segmentation masks and generated instructions), and a predefined set of potential error conditions (\eg, “the described intent matches multiple objects” or “the intent is not sufficiently specific”). The MLLM evaluates each case and returns one of three outcomes: “Yes” (non-compliant, to be discarded), “No” (valid), or “Uncertain.” All samples marked as “No” or “Uncertain” proceed to a manual curation phase. Five expert annotators review these cases, referencing the reasoning outputs from the verification stage. Each annotation is cross-checked against rigorous criteria, such as safety, focus relevance, unique object correspondence, and contextual consistency to guarantee the overall quality of the final dataset.

\subsubsection{Data Statistics}

We apply the above annotation pipeline to $4$ existing datasets: DUTS~\cite{duts}, LVIS~\cite{lvis}, PACO-LVIS~\cite{paco}, and EgoObjects~\cite{egoobjects}. The OC-SODBench dataset comprises 33k images with 152k high-quality instruction-mask pairs, which is further divided into training ($133.9$K), validation ($7.4$K), and test ($11$K) sets. The test set is further split into three subsets: free-viewing ($1.3$K), preference-driven ($5.3$K), and intent-driven ($4.3$K).
Beyond its large scale, the benchmark is characterized by its diversity and complexity, as depicted in \cref{fig:statistics}. The word clouds in \cref{fig:statistics}(a) highlight the prevalent object and part category, which illustrate the rich diversity of target objects, user intentions and preferences. 
This scene complexity is further quantified in \cref{fig:statistics}(b), where a histogram of object counts per image reveals a challenging distribution of potential targets. On average, there are $2.23$  objects with masks per image, with the maximum number of objects in a image being $15$. This promotes robust multi-target reasoning. 
Finally, \cref{fig:statistics}(c) presents a pie chart of the target mask's pixel-area ratio. While filtering out targets with low-percentage masks, we also retain targets with a smaller ratio but meaningful focus, in order to enhance the richness of the data and increase the focusing difficulty of the significant targets. The wide distribution of scales, particularly the significant portion of low-percentage masks (accounting for less than $5$\% of the image area), ensures the benchmark rigorously evaluates a model's ability to localize targets of varying sizes, especially its capability to focus on smaller, salient objects.

%% file: sec_arxiv/4_method.tex
\label{sec:method}

\begin{figure*}[t]

  \centering
    \includegraphics[width=0.8\linewidth]{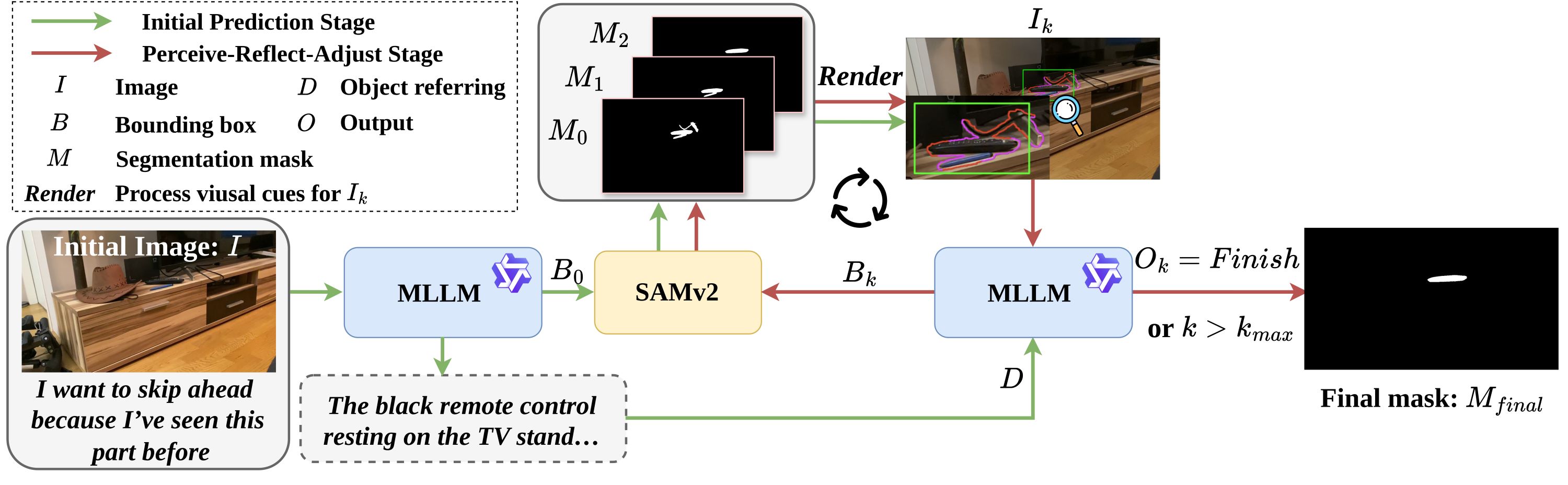}
    \vspace{-3mm}
        \caption{Pipeline of the proposed OC-SODAgent. 
        In the initial prediction stage (the green arrows), given the input image with the instruction, the MLLM first parses the user's intent/preference and generates an initial bounding box $B_0$, which is then processed by SAMv2 to produce an initial mask $M_0$. Based on the predicted bounding box and mask, a rendered image is synthesized, overlaying the box and the contour of the region of interest. The process then enters the Perceive–Reflect–Adjust cycle (the red arrows), where the MLLM together with SAMv2 iteratively perceive, reason, and refine the output, repeating until convergence to produce the final result $M_{final}$.
        }
  \label{fig:method}

\end{figure*}

\section{Agentic Baseline for OC-SOD}
We propose OC-SODAgent, an agentic framework for observer-centric salient object detection. As illustrated in \cref{fig:method}, OC-SODAgent consists of two key components: an MLLM and a segmentation model SAMv2 \cite{sam2}. It operates in two stages: an initial prediction stage followed by an Perceive–Reflect–Adjust cycle, progressively improving saliency prediction accuracy based on contextual reasoning and feedback.

\subsection{Initial Prediction}
During the initial prediction, the MLLM is prompted to understand the instruction, perceive the input image, and reasons about the salient regions accordingly. The pixel-level segmentation mask is then achieved by calling the SAMv2 \cite{sam2} model. Specifically, given the input image $I$ and current task instruction $T$, the MLLM $\mathcal{M}$ proceeds as follows: 

\begin{equation}
(B_0, D) = \mathcal{M}(I, T),
\label{eq:1}
\end{equation}
where $B_0 = \{b_i\}_{i=1}^{N}$ denotes a set of $N$ predicted bounding boxes that tightly enclosed each potential salient object regions, and $D$ represents the referring descriptions and explanations for the corresponding objects. Subsequently, the bounding boxes $B_0$ serving as box prompts are fed into SAMv2 $\mathcal{S}$ along with the input image $I$ to generate the initial segmentation masks $M_0$. The resulting masks $M_0$ and the referring descriptions $D$ then serve as  input for the subsequent refinement process.

\subsection{Perceive-Reflect-Adjust}
The Perceive-Reflect-Adjust stage is designed to iteratively refine the segmentation accuracy by taking advantage of the strong reasoning and reflection power of the MLLM-based agent. \cref{alg:frm} demonstrates the detailed pseudo-code of this process.

During the k-th iteration ($k\ge 1$), the segmentation masks $M_{k-1}$ from the previous step is first converted into a visual prompt for MLLM to perceive. To alleviate the impact caused by object occlusion, we extract 
the fine-grained object contours $\partial M_{k-1}$ from the masks $M_{k-1}$. This information and $M_{B-1}$ are further rendered onto the original image $I$ to create an enhanced visual input $I_k$ (See \cref{fig:method} as an example).
Subsequently, the MLLM $\mathcal{M}$ receives $I_k$, the referring description $D$, and a pre-defined reflection prompt $T_{r}$ to perform reflection:
\begin{equation}
B_k, O_k  = \mathcal{M}(I_k, D, T_{r}),
\label{eq:4}
\end{equation}
where the prompt $T_{r}$ specifies the conditions for termination and guides the MLLM to reflect how to adjust the bounding boxes to more accurately locate the salient objects; the output $B_k$ denotes updated bounding boxes; and $O_k$ represents an indicator of whether the Perceive-Reflect-Adjust process could be terminated.

Given the updated bounding boxes, SAMv2 model is further invoked to produce a set of refined segmentation masks $M_k$. The above process if iteratively proceeds  until the iteration count $k$ exceeds the maximum iteration number or the MLLM output a termination signal ($O_k=\text{Finish}$).

With well-designed input prompt, OC-SODAgent can deliver superior performance on OC-SODBench even without fine-tuning. With further fine-tuning (as shown in \cref{sec:experiment}), additional performance boost is achieved. We provide detailed input prompt in the supplementary materials. 

\renewcommand{\algorithmicrequire}{\textbf{Input:}}
\renewcommand{\algorithmicensure}{\textbf{Output:}}
\begin{algorithm}[t]
\footnotesize
\caption{Perceive-Reflect-Adjust Mechanism}
\label{alg:frm}
\scalebox{0.9}{
\begin{minipage}{1.1\linewidth} 
\begin{algorithmic}[1]
\REQUIRE Zero-shot segmentation model $\mathcal{S}$, MLLM (reflection agent) $\mathcal{M}$, original image $I$, referring description $D$, the k-th interaction mask $M_{k}$, the k-th interaction bounding box $B_k$, reflection prompt $T_{r}$, maximum iteration $K_{max}$.
\ENSURE Final refined mask $M_{final}$.
\STATE Initialize iteration $k \leftarrow 1$.
\FOR{$k=1,2,\ldots,K$}
    \STATE $\partial M_{k-1} \leftarrow \texttt{Process}(M_{k-1})$ \hfill // extract visual cue
    \STATE $I_k \leftarrow \texttt{Render}(I, \partial M_{k-1}, B_{k-1})$ \hfill // render focused region
    \STATE $ B_k, O_k \leftarrow \mathcal{M}(I_k, D, T_{r})$ \hfill // MLLM reflection
    \IF{$O_k = \texttt{Finish}$}
        \STATE $M_{final} \leftarrow M_{k-1}$; \textbf{break}
    \ENDIF
    \STATE $M_k \leftarrow \mathcal{S}(I, B_k)$ \hfill // re-segment
    \STATE $k \leftarrow k + 1$
\ENDFOR
\RETURN $M_{final}$
\end{algorithmic}
\end{minipage}
}
\end{algorithm}

%% file: sec_arxiv/5_experiment.tex
\section{Experiments}
\label{sec:experiment}

\begin{table}[t]
\tiny
\renewcommand{\arraystretch}{1.0}
\centering
\caption{Quantitative evaluation on the \textbf{free-viewing} mode of OC-SODBench test dataset.
}
\label{tab:free-viewing_result}
\resizebox{1.0\linewidth}{!}{
\begin{tabular}{l | ccccc} 
\toprule
\multirow{2}{*}{\textbf{Method}} & \multicolumn{5}{c}{\textbf{Free-Viewing}} \\
\cline{2-6}
& gIoU  & cIoU  & S$_{m}$  & F$_{m}$  & E$_{m}$  \\
\midrule
\multicolumn{6}{l}{\textit{Traditional SOD}} \\ 
\ VST \cite{vst} & 81.28 & 83.96 & 92.82 & 92.34 & 97.05 \\
\ ICON \cite{icon}& 83.07 & 86.36 & 93.81 & 93.74 & 97.92 \\
\ SelfReformer \cite{selfreformer}& 82.56 & 85.62 & 93.59 & 93.52 & 97.44 \\
\ VSCode \cite{vscode}& 85.33 & 87.77 & 94.94 & 95.09 & 98.23 \\
\ MDSAM \cite{mdsam}& 84.38 & 85.90 & 94.20 & 94.50 & 97.51 \\
\ FOCUS \cite{focus}& 85.62 & 88.36 & 94.55 & 94.41 & \textbf{98.33} \\
\midrule
\multicolumn{6}{l}{\textit{MLLMs in training-free setting}} \\ 
\ MMR-7B \cite{mmr}& 57.98 & 42.79 & 71.51 & 62.41 & 77.56 \\
\ PixelLM \cite{pixellm}& 76.25 & 78.95 & 88.85 & 86.69 & 94.50 \\
\ LISA-7B \cite{lisa}& 81.80 & 80.38 & 89.57 & 87.14 & 94.16 \\
\textbf{OC-SODAgent} & 87.04 & 86.76 & 94.74 & 94.55 & 96.69 \\

\midrule
\multicolumn{6}{l}{\textit{MLLMs with finetuning}} \\ 
\ LISA-7B$_{\text{FT}}$ & 88.71 & 88.91 & 92.44 & 92.58 & 96.70 \\
\textbf{OC-SODAgent$_{\text{FT}}$} & \textbf{89.13} & \textbf{88.92} & \textbf{95.84} & \textbf{95.75} & 98.02 \\
\bottomrule
\end{tabular}
}
\end{table}
\begin{table*}[t]
\renewcommand{\arraystretch}{1.0}
\centering
\footnotesize  
\caption{Quantitative evaluation on the \textbf{intent-driven} and \textbf{preference-driven} modes of OC-SODBench test dataset.
}
\label{tab:complexity_guided}
\resizebox{0.8\linewidth}{!}{
\begin{tabular}{l | ccccc | ccccc}
\toprule

\multirow{2}{*}{\textbf{Method}} & \multicolumn{5}{c|}{\textbf{Intent-Driven}} & \multicolumn{5}{c}{\textbf{Preference-Driven}} \\
\cline{2-11} 
& gIoU  & cIoU  & S$_{m}$  & F$_{m}$  & E$_{m}$  & gIoU  & cIoU  & S$_{m}$  & F$_{m}$  & E$_{m}$  \\
\midrule

\multicolumn{11}{l}{\textit{MLLMs in training-free setting}} \\
\ MMR-7B \cite{mmr}& 11.58 & 11.28 & 47.03 & 14.58 & 56.22 & 19.08 & 15.54 & 46.24 & 24.87 & 57.70 \\
\ PixelLM \cite{pixellm}& 21.67 & 23.74 & 55.03 & 26.90 & 59.76  & 44.42 & 41.12 & 67.69 & 56.36 & 71.09 \\
\ LISA-7B \cite{lisa}& 14.53 & 14.83 & 33.68 & 10.86 & 30.90 & 33.02 & 25.49 & 52.28 & 37.66 & 54.73 \\
\textbf{OC-SODAgent} & 26.23 & 28.83 & 62.90 & 39.97 & 69.23 & 60.61 & 60.54 & 81.08 & 73.97 & 85.79 \\

\midrule
\multicolumn{11}{l}{\textit{MLLMs with finetuning}} \\
\ LISA-7B$_{\text{FT}}$ & 27.18 & \textbf{33.09} & 56.60 & 25.92 & 63.52 & 59.66 & 42.90 & 75.19 & 66.72 & 78.97 \\
\textbf{OC-SODAgent$_{\text{FT}}$} & \textbf{29.86} & 32.17 & \textbf{65.28} & \textbf{41.21} & \textbf{72.56} & \textbf{62.82} & \textbf{63.99} & \textbf{83.10} & \textbf{75.53} & \textbf{87.69} \\
\bottomrule
\end{tabular}
}
\end{table*}


\subsection{Implementation Details}

To ensure high-quality training data, our data pipeline leverages the Qwen3-VL-235B-A22B-Instruct and Qwen3-VL-235B-A22B-Thinking models for data generation and filtering. During generation, we employ a $top\_p$ of $0.95$ and a $temperature$ of $0.8$ to enhance diversity, while filtering adheres to the official default hyperparameters. 
Our proposed OC-SODAgent is built upon the Qwen3-VL-8B-Instruct model as its reasoning core, integrated with the official pre-trained SAM2-Hiera-Large as the segmentor. 



For a fair comparison, we fine-tune both OC-SODAgent and LISA-7B on the OC-SODBench training set using identical configurations. We use the Adam optimizer with a learning rate of $3e-7$, a per-device batch size of $2$, and gradient accumulation steps of $2$. A cosine learning rate scheduler is employed with a warmup ratio of $0.03$. We set weight decay to $0$ and apply gradient clipping with a max norm of $1$. We utilize mixed-precision training to enhance efficiency. All other hyperparameters follow the respective official implementations. The training and evaluation are conducted on 8 NVIDIA RTX 4090 GPUs.


\subsection{Evaluation Metrics}

We evaluate the models using a combination of saliency and segmentation metrics to ensure a comprehensive evaluation of our observer-centric saliency detection paradigm.
For saliency assessment, we employ the Structure-measure ($S_m$)~\cite{s-measure}, maximum F-measure ($F_m$), and Enhanced-alignment measure ($E_m$)~\cite{e-measure} to quantify detection performance. For segmentation quality, we report generalized IoU (gIoU) for per-instance accuracy and cumulative IoU (cIoU) for dataset-level consistency following~\cite{lisa,gres}.



\subsection{Overall Comparsion}

We conduct comprehensive comparisons on the OC-SODBench test dataset under three distinct modes: free-viewing, intent-driven, and preference-driven. 
For the free-viewing mode, we compare against six state-of-the-art SOD methods (VST~\cite{vst}, ICON~\cite{icon}, SelfReformer~\cite{selfreformer}, VSCode~\cite{vscode}, MDSAM~\cite{mdsam}, and FOCUS~\cite{focus}) along with three MLLM-based reasoning segmentation approaches (MMR-7B~\cite{mmr}, PixelLM~\cite{pixellm}, and LISA-7B~\cite{lisa}). 
For the intent-driven and preference-driven modes, comparisons are performed exclusively against the three MLLM-based reasoning segmentation models. 
For all MLLM-based methods, we first evaluate their performance in a training-free manner to assess their zero-shot generalization.
Additionally, we finetune these methods on the OC-SODBench training dataset, and report the results of our OC-SODAgent with the representative method LISA-7B~\cite{lisa}. More fine-tuned results can be found in the supplementary materials.



\textbf{Free-viewing Mode.}
As shown in Table~\ref{tab:free-viewing_result}, the specialized SOD models perform well under the free-viewing mode, which aligns with the conventional SOD paradigm. Despite this, our OC-SODAgent yields competitive results without any training. After fine-tuning, the MLLM-based methods LISA-7B and OC-SODAgent are further boosted, with ours ultimately achieving superior performance.

\textbf{Intent-driven and Preference-driven Modes.} 
Table~\ref{tab:complexity_guided} provides the quantitative results for both intent-driven and preference-driven modes\footnote{It is important to emphasize that in the intent-driven mode, user requests may focus on fine-grained objects or even part-level regions (e.g., a mouse, button), making this task notably more challenging than the other two modes.}, where saliency is conditioned on observer's intent or long-term preference.
Under the training-free setting, our OC-SODAgent achieves the highest scores across all metrics for both modes, showcasing the superior zero-shot capability among the compared methods. After fine-tuning, both LISA and our OC-SODAgent exhibit significant performance gains, with LISA showing a particularly notable improvement. Despite these improvements, our fine-tuned OC-SODAgent still maintains a significant advantage. 
Moreover, the performance enhancement observed after fine-tuning validates the effectiveness and practical utility of the proposed OC-SODBench dataset, which we hope will provide a new paradigm for observer-centric salient object detection research. 

\begin{figure*}[t]
  \centering
    \includegraphics[width=0.95\linewidth]{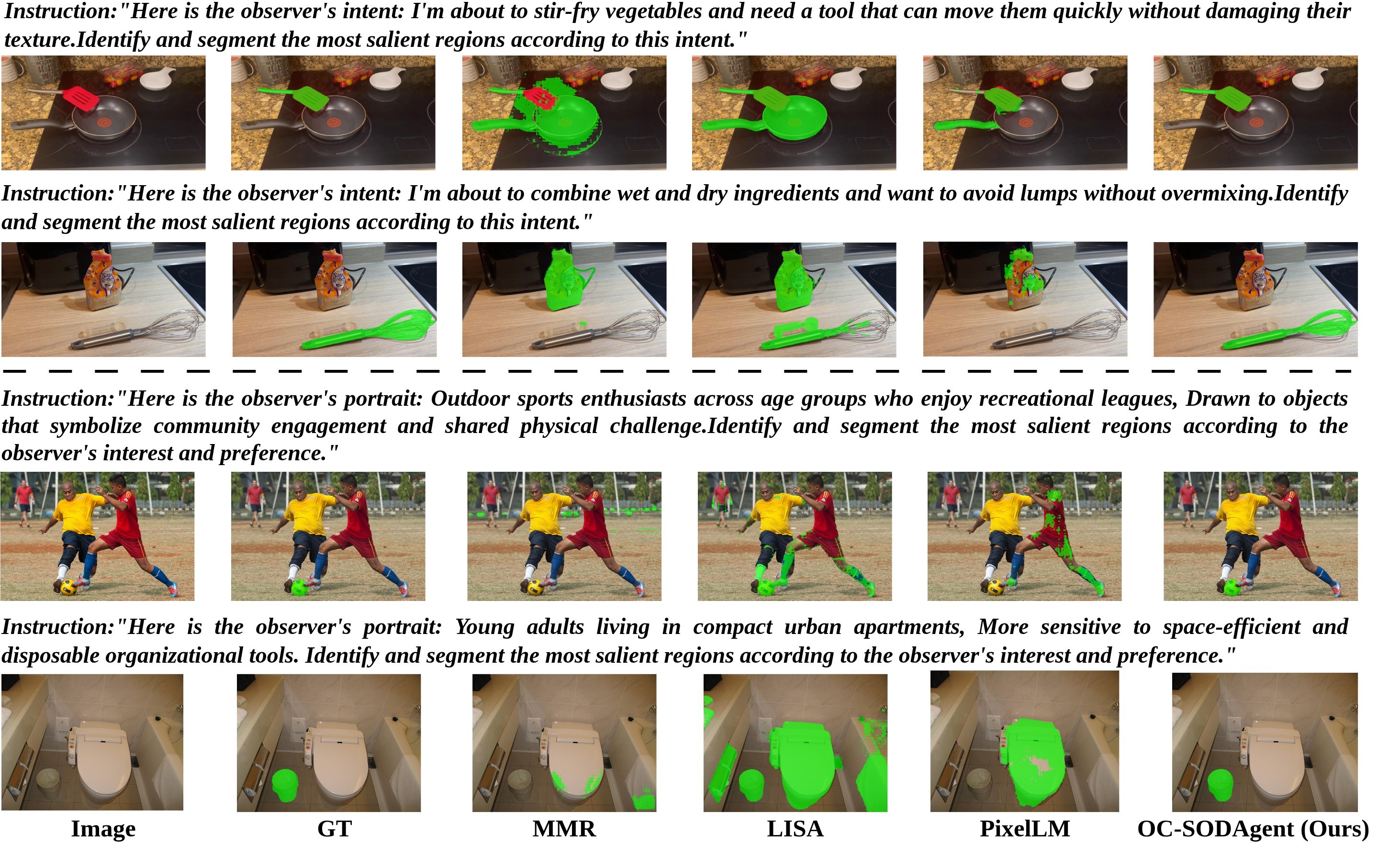}
        \caption{Example of visualization results under the intent-driven mode (the top two rows) and preference-driven mode (the last two rows). }
  \label{fig:visual}
\end{figure*}

\textbf{Visual Comparison.} Figure~\ref{fig:visual} visualizes the SOD results for both the intent-driven and preference-driven modes. It shows that the proposed OC-SODAgent can accurately identify the salient objects in the complex scenes, which well align with the diverse interests and intents of different observers. More visualization results can be found in the supplementary materials.




\subsection{Ablation Study}
To further investigate the impact of key components in our approach, we conduct ablation studies under the intent-driven mode, which presents a more challenging scenario. For efficient validation, we select a subset of 100 images randomly sampled at regular intervals from the intent-driven subset, with up to five objects considered per image.

\textbf{Effectiveness of Perceive-Reflect-Adjust Mechanism.}
We investigate the efficacy of the Perceive-Reflect-Adjust mechanism in OC-SODAgent and the impact of its iteration number $k$ on the final performance. 
Therefore, we increase the value of $k$ from $1$ to $5$, and summarize the results in Table~\ref{tab:ablation_reflection}.
Compared to the baseline ($k=1$) that denotes a single forward pass without iterative reflection, introducing one more iteration ($k=2$) yields a substantial performance gain, indicating that incorporating a feedback and refinement strategy can effectively improve the segmentation accuracy. 
Moreover, while increasing the number of iterations can further enhance performance, we observe minimal gains beyond three rounds. Considering that additional iterations also increase inference time, we ultimately set $k=3$ as the optimal trade-off between accuracy and efficiency.

\begin{table}[t]

\renewcommand{\arraystretch}{1.0}
\centering
\caption{
Ablation study on the perceive-reflect-adjust mechanism.
}
\label{tab:ablation_reflection}
\resizebox{0.6\linewidth}{!}{%
\begin{tabular}{l | cc}
\toprule
Iteration Number & gIoU  & cIoU  \\
\midrule
$k$ = 1 (w/o reflection) & 34.81 & 37.87 \\
$k$ = 2  & 36.46 & 39.61 \\
\textbf{$k$ = 3  (Optimal)} & \textbf{36.97} & \textbf{40.24} \\
$k$ = 4  & 37.04 & 40.30 \\
$k$ = 5  & 37.12 & 40.37 \\
\bottomrule
\end{tabular}%
}
\end{table}

\textbf{Different MLLM Backbones.} 
We ablate the MLLM backbone in our OC-SODAgent to assess its effect on final performance.
We compare the Qwen3-VL-8B-Instruct (default choice) with Qwen3-VL-32B-Instruct and Qwen3-VL-30B-A3B-Instruct as the backbone under otherwise identical settings. Table~\ref{tab:mllm_ablation} shows the generality of our framework, where stronger backbones yield improved performance, suggesting our OC-SODAgent framework can benefit from future advances in foundation models.




\begin{table}[t]
\renewcommand{\arraystretch}{1.2}
\centering
\caption{Ablation study on MLLM backbones for OC-SODAgent. 
}
\label{tab:mllm_ablation}
\resizebox{0.75\linewidth}{!}{%
\begin{tabular}{l | cc}
\toprule
Backbone & gIoU & cIoU \\
\midrule
qwen3-vl-8b-instruct (Default) & 36.97 & 40.24 \\
qwen3-vl-32b-instruct          & 42.66 & 44.85 \\
qwen3-vl-30b-a3b-instruct      & \textbf{47.07} & \textbf{48.76} \\
\bottomrule
\end{tabular}%
}
\end{table}

%% file: sec_arxiv/6_conclusion.tex
\section{Conclusion}
\label{sec:conclusion}

This work revisits the long-standing ill-posed problem of Salient Object Detection (SOD) from an observer-centric perspective. We propose Observer-Centric SOD (OC-SOD), which explicitly models how observers define salient objects under various influencing factors. To this end, we develop OC-SODBench, the first large-scale observer-centric SOD benchmark encompassing three modes, including free-viewing, intent-driven, and preference-driven. Furthermore, an agentic baseline method OC-SODAgent is designed, which mimics the human ``perceive–reflect–adjust'' reasoning via multi-step reflection.
Our work lays a foundation for personalized, context-aware saliency modeling, opening avenues for more human-aligned and flexible visual understanding in real-world applications.





\newpage

%% file: sec_arxiv/X_suppl.tex
\clearpage
\setcounter{page}{1}
\maketitlesupplementary


\section*{Appendix A. Prompts for Data Annotation and Observation Modes}
\addcontentsline{toc}{section}{Appendix A. Prompts for Data Annotation and Observation Modes}

\begin{figure*}[h]
  \centering
  \includegraphics[width=0.75\linewidth]{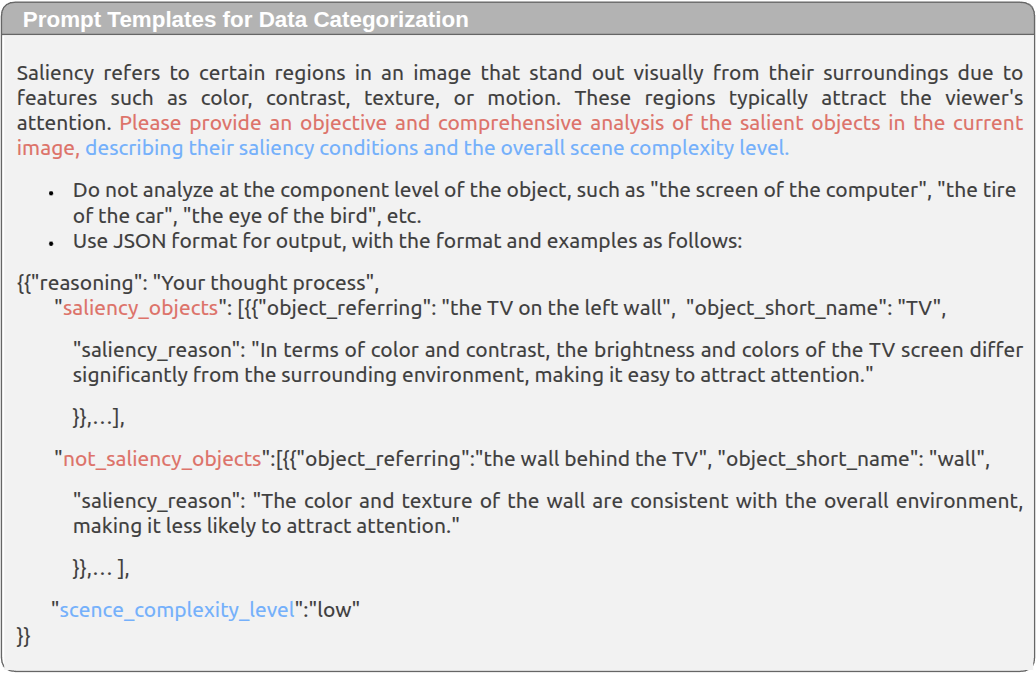}
  
  \caption{The template instructs the MLLM to perform an objective saliency analysis by identifying salient and non-salient objects, providing reasoning for each decision, and assessing the overall scene complexity.}
  \label{fig:data_cat}
\end{figure*}

\begin{figure*}[h]
  \centering
  \includegraphics[width=0.75\linewidth]{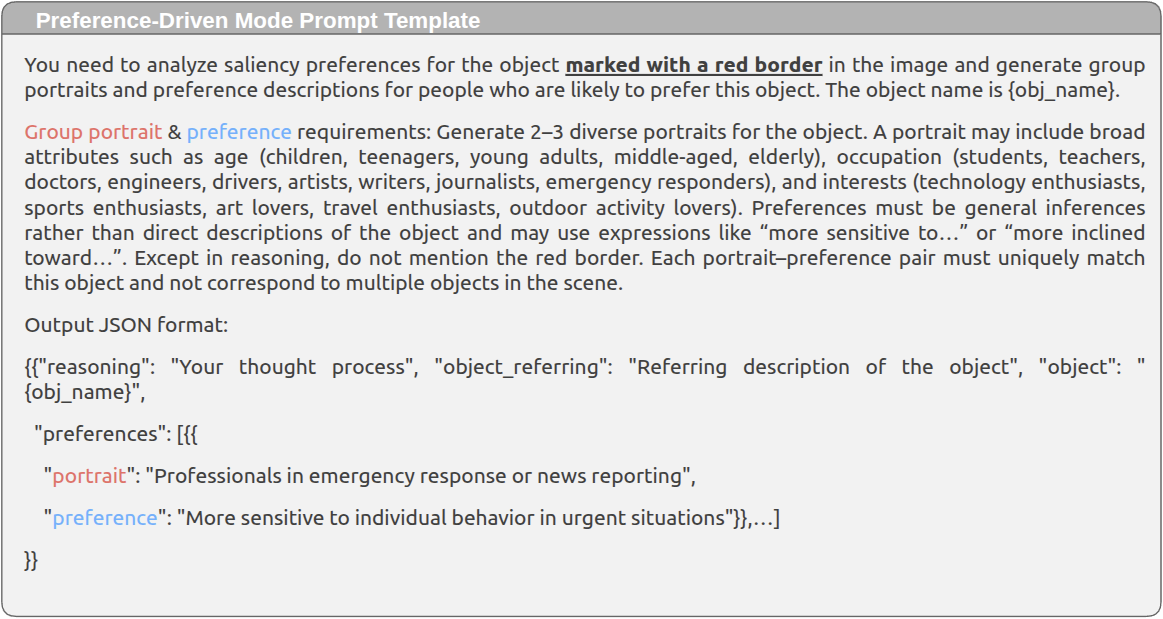}
  
  \caption{The template guides the MLLM to generate group portraits and preference descriptions by analyzing the saliency preferences for a red-bordered object, outputting the reasoning, object description, and 2-3 diverse portrait-preference pairs in JSON format.}
  \label{fig:preference_gen}
\end{figure*}

\begin{figure*}[h]
  \centering
  \includegraphics[width=0.75\linewidth]{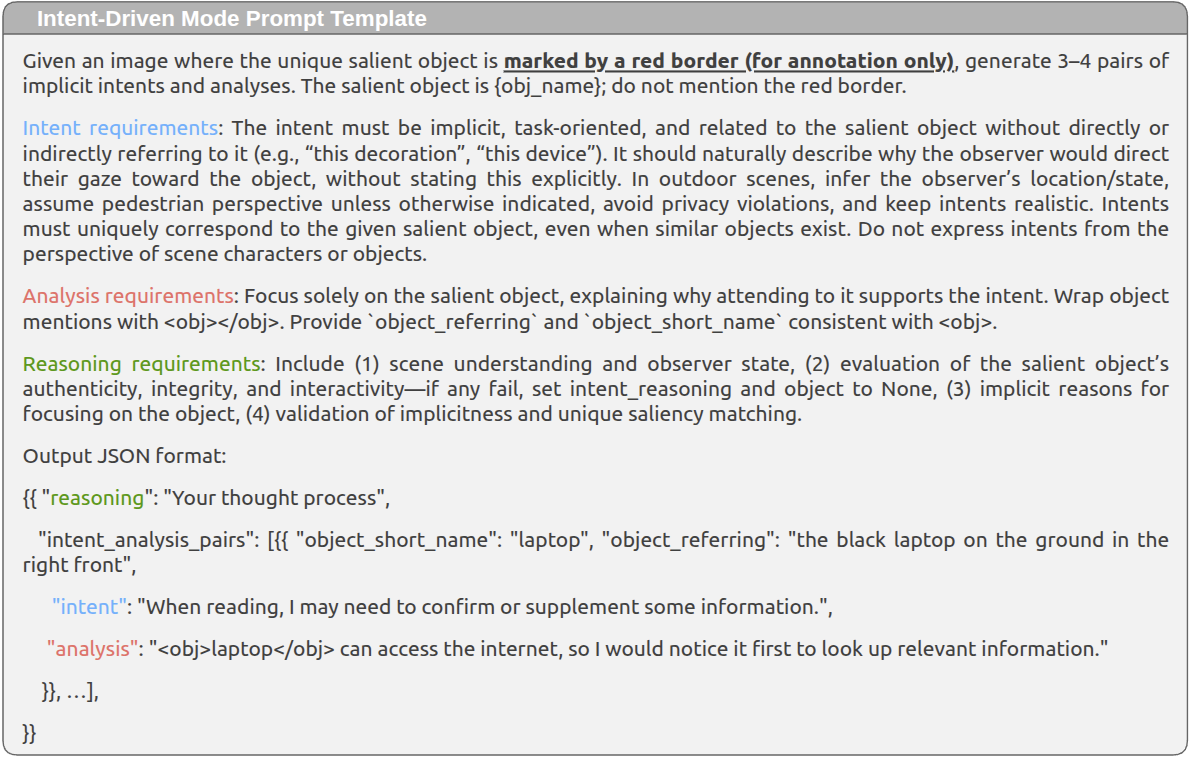}
  
  \caption{The template guides the MLLM to generate implicit intents and analyses by formulating task-oriented goals for a red-bordered object, outputting the reasoning, object description, and 3-4 intent-analysis pairs in JSON format.}
  \label{fig:intent_gen}
\end{figure*}

\subsection*{A.1 Prompt Templates for Data Categorization}
\label{app:A1}
 This subsection provides the prompt templates used in the data categorization stage of the MLLM-driven annotation pipeline. These templates correspond to the \textbf{Data Categorization} step described in Sec.~3.2.1 of the main paper. 

\paragraph{Templates}
As shown in \cref{fig:data_cat}, our data categorization prompt contains two major components, both adapted from the full prompt template illustrated in the supplementary figure.

\begin{itemize}[leftmargin=12pt]

\item \texttt{Object saliency analysis.}
This part instructs the model to perform an \emph{objective and comprehensive analysis} of salient objects in the image.  
It emphasizes:
\begin{itemize}[leftmargin=10pt]
  \item identifying visually prominent objects based on features such as color, contrast, texture, and motion,
  \item avoiding component-level analysis (e.g., ``the screen of the laptop'', ``the tire of the car''),
  \item Output the reasoning and saliency judgment strictly in \textbf{JSON} format.
\end{itemize}

\item \texttt{Scene complexity assertion.}
This part asks the model to classify the scene difficulty and provides a summary-level judgment used later for observation-mode assignment.  
It requires the MLLM to output:
\begin{itemize}[leftmargin=10pt]
  \item \texttt{saliency\_objects}: salient targets with reason,
  \item \texttt{not\_saliency\_objects}: non-salient targets with reason,
  \item \texttt{scene\_complexity\_level}: \{``low'', ``medium'', ``high''\}.
\end{itemize}
\end{itemize}

\subsection*{A.2 Prompt Templates for Data Generation in Observation Modes}
\label{app:A2}
 These prompt templates are the data-generation prompts referenced in Sec.~3.1 (``The OC-SOD Task Setting'') and Sec.~3.2.1 (``Instruction Generation'') of the main paper. Each template is intended to produce one instruction (textual prompt) that, together with a mask, forms an \emph{instruction--mask} pair in OC-SODBench.

\subsubsection*{A.2.1 Free-Viewing Mode Data Source} Corresponding to Sec.~3.2.1 (Free-viewing mode will be annotated with a fixed instruction ...'').\\ \noindent\textbf{Data Source Description:} This mode directly corresponds to the results from the \textbf{Data Categorization} step where the \texttt{scene\_complexity\_level} was determined to be \textbf{``low''}. The data for this mode, including the target salient objects and the saliency reasoning, are directly utilized from the output of that analysis.

\subsubsection*{A.2.2 Preference-Driven Mode Prompt Template}
Corresponding to Sec.~3.2.1 (preference-conditioned instruction construction).

\paragraph{Template}
As illustrated in \cref{fig:preference_gen}, the preference-driven mode prompt instructs the MLLM to analyze saliency preferences for the unique salient object marked by a red border (for annotation only) and to generate corresponding group portraits and preference descriptions. The prompt contains the following major components.

\begin{itemize}[leftmargin=12pt]

\item \texttt{Group portrait \& preference requirements.}
the model must generate 2-3 diverse portraits and preference pairs for the given salient object.
These pairs should:
\begin{itemize}[leftmargin=10pt]
\item Include broad attributes for the portrait, such as age (e.g., young adults, middle-aged, elderly) or occupation/interests (e.g., students, teachers, artists, emergency responders, outdoor activity lovers).
\item Define preferences as general inferences (e.g., ``more inclined toward...'') rather than direct descriptions of the object.
\item Uniquely match the specified salient object and not correspond to multiple objects in the scene.
\item Avoid mentioning the red border (used for annotation) in the reasoning or description.
\end{itemize}

\end{itemize}

\paragraph{Output format}
The model outputs 2–3 portrait–preference pairs in JSON format. The output includes \texttt{reasoning}, \texttt{object\_referring}, \texttt{object}, and a \texttt{preferences} list, following the structure shown in \cref{fig:preference_gen}.

\begin{figure*}[h]
  \centering
  \includegraphics[width=0.75\linewidth]{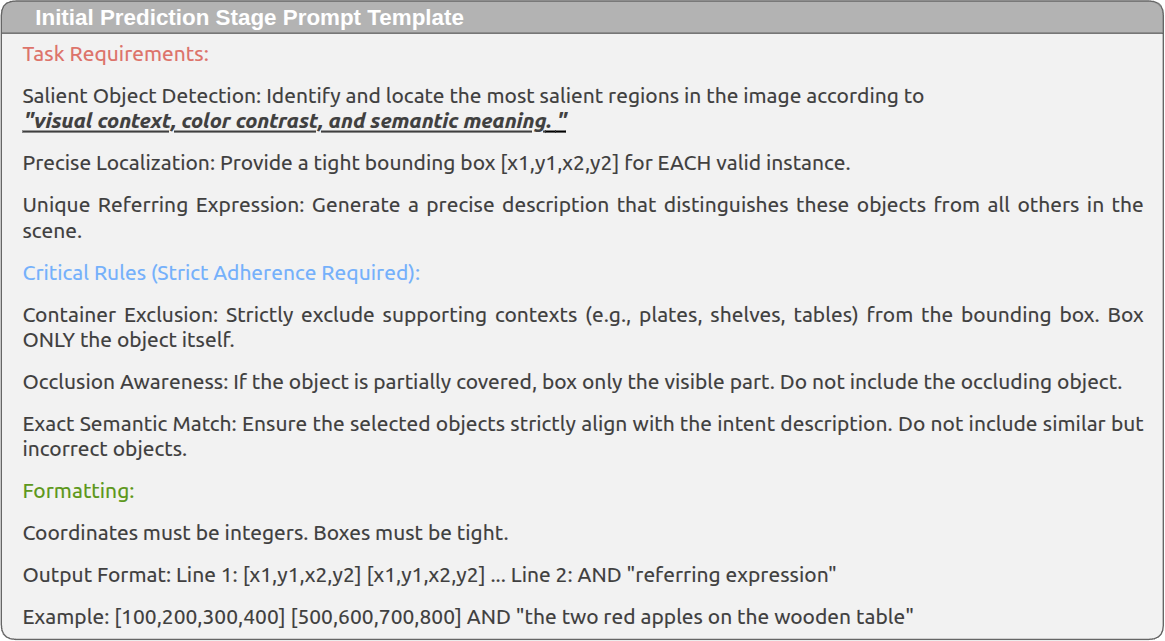}
  
  \caption{The prompt template for the Initial Prediction Stage of the OC-SODAgent. The task description dynamically adapts to the specific observation mode: it utilizes a general Salient Object Detection description for Free-Viewing mode (focusing on visual prominence), while enforcing alignment with specific Observer-Centric subjective intents or preferences for the Intent-Driven and Preference-Driven modes.}
  \label{fig:agent_stage1}
\end{figure*}

\begin{figure*}[h]
  \centering
  \includegraphics[width=0.75\linewidth]{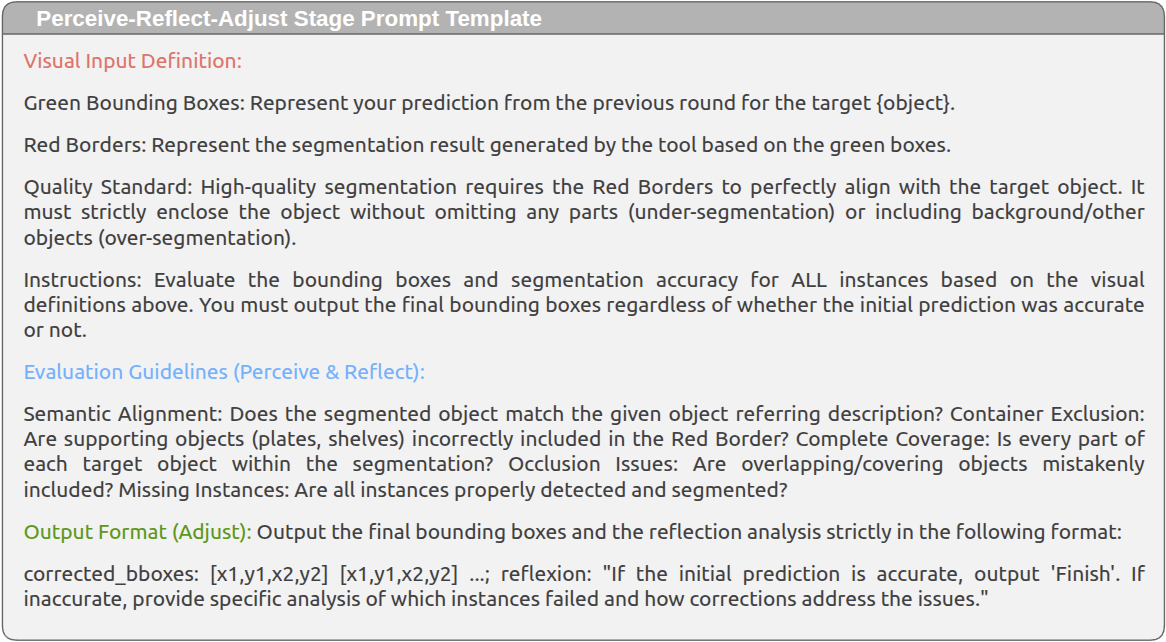}
  
  \caption{The prompt template for the Perceive--Reflect--Adjust Stage of the OC-SODAgent. It provides the agent with explicit visual definitions (Green Boxes for previous predictions, Red Borders for segmentation results) and evaluation guidelines (e.g., container exclusion, occlusion handling). The agent uses this prompt to assess segmentation quality and generate corrected bounding boxes if necessary.}
  \label{fig:agent_stage2}
\end{figure*}

\subsubsection*{A.2.3 Intent-Driven Mode Prompt Template}
Corresponding to Sec.~3.2.1 (intent-conditioned instruction formulation).

\paragraph{Template}
As illustrated in \cref{fig:intent_gen}, the intent-driven mode prompt instructs the MLLM to construct implicit task-oriented intents and corresponding analyses for the unique salient object marked by a red border (for annotation only). The prompt contains three major components.

\begin{itemize}[leftmargin=12pt]

\item \texttt{Intent requirements.}
The model must generate implicit, task-driven intents related to the given salient object while avoiding any direct or indirect reference to the object itself (e.g., ``this decoration'', ``this device'').  
These intents should:
\begin{itemize}[leftmargin=10pt]
    \item reflect plausible motivations for why the observer would direct their gaze toward the object;
    \item remain realistic and context-aware, especially for outdoor scenes (assuming a pedestrian perspective unless otherwise implied);
    \item avoid violating privacy or attributing intentions to characters or objects in the image;
    \item uniquely correspond to the given salient object, even in the presence of similar objects.
\end{itemize}

\item \texttt{Analysis requirements.}
The analysis must focus exclusively on the salient object and explain why attention to this object supports the generated intent.  
The prompt enforces:
\begin{itemize}[leftmargin=10pt]
    \item using \verb|<obj></obj>| tags around the object name;
    \item including both \texttt{object\_referring} and \texttt{object\_short\_name} consistent with the tagged object;
    \item ensuring the analysis directly relates to the intent and cannot map to multiple objects.
\end{itemize}

\item \texttt{Reasoning requirements.}
The model must explicitly document its thought process by:
\begin{itemize}[leftmargin=10pt]
    \item summarizing the scene and identifying the observer’s likely state,
    \item assessing the salient object’s authenticity, integrity, and interactivity—if any condition fails, the intent and object fields are set to \texttt{None};
    \item generating implicit, diverse reasons that distinguish this object from others in the scene;
    \item validating implicitness and unique saliency matching of each generated intent.
\end{itemize}

\end{itemize}

\paragraph{Output format}
The model outputs 3–4 intent–analysis pairs in JSON format, following the structure shown in \cref{fig:intent_gen}.

\subsection*{A.3 Instruction Examples Generated for Each Observation Mode}
\label{app:A3}
 Example \textbf{Instructions} (i.e., the generated textual prompts that serve as the dataset labels) for each observation mode. These reflect the \emph{instruction--mask} pairs discussed in Sec.~3.2 (``The OC-SODBench Dataset'').

\subsubsection*{A.3.1 Instruction Example: Free-Viewing Mode}
\begin{quote}
\texttt{``Identify and segment the most salient regions according to visual context, color contrast, and semantic meaning.''}
\end{quote}

\subsubsection*{A.3.2 Instruction Example: Preference-Driven Mode}
\begin{quote}
\texttt{``Here is the observer's portrait: \{A foodie who loves freshly baked goods and flavorsome bread\}. Identify and segment the most salient regions according to the observer's interest and preference.''}
\end{quote}

\subsubsection*{A.3.3 Instruction Example: Intent-Driven Mode}
\begin{quote}
\texttt{``Here is the observer's intent: \{I want to check and reply to my email\}. Identify and segment the most salient regions according to this intent.''}
\end{quote}


\begin{figure*}[h]
  \centering
  \includegraphics[width=1\linewidth]{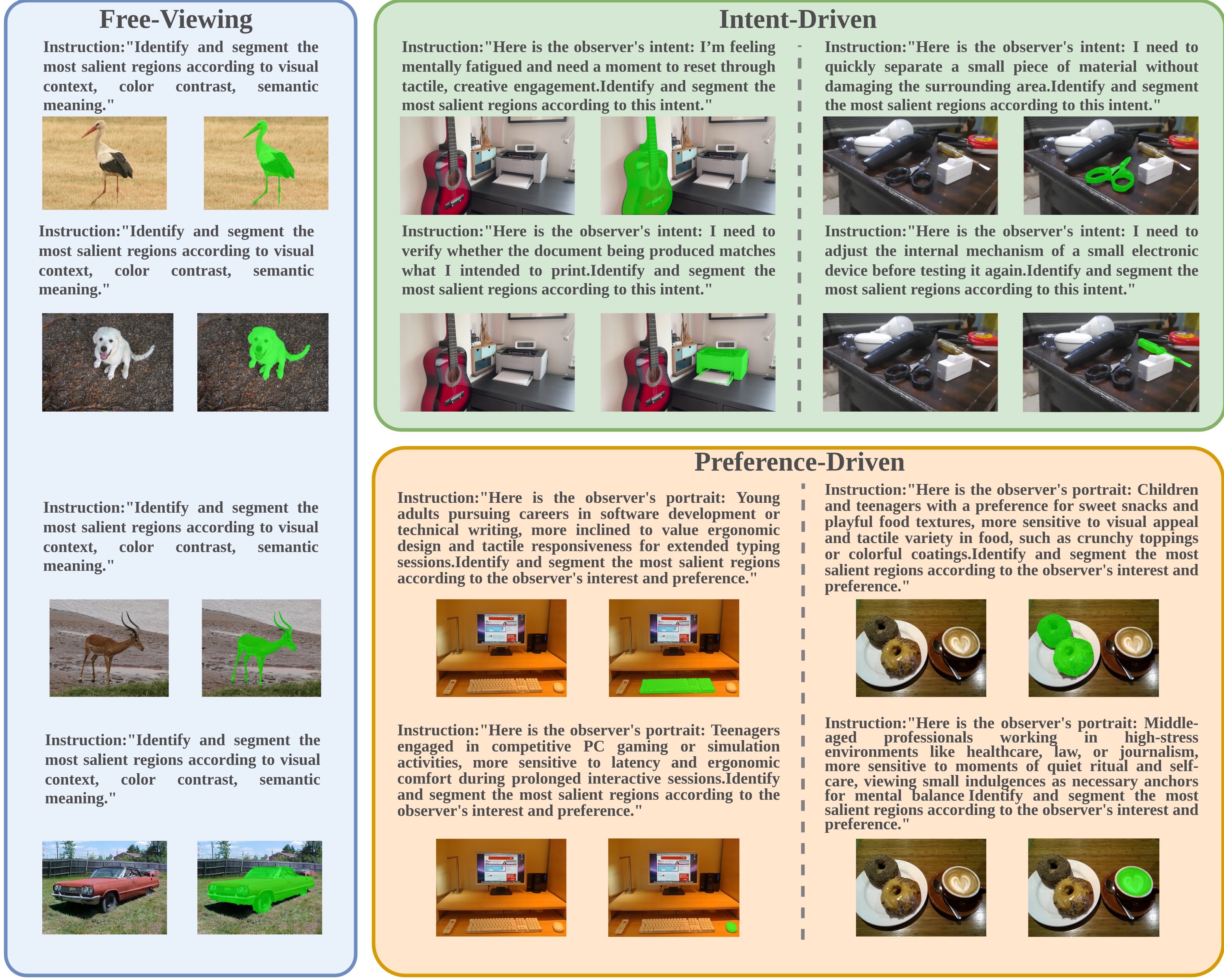}
  \caption{\textbf{Representative examples from the OD-SODBench dataset illustrating the three distinct Observer-Centric modes.} The figure demonstrates how the model adapts its segmentation targets based on different instruction types for the same visual input: \textbf{(Left) Free-Viewing}, where the model identifies regions based on inherent visual saliency and semantic meaning; \textbf{(Top Right) Intent-Driven}, where segmentation is guided by specific user tasks (e.g., distinguishing a guitar for creative engagement versus a printer for document verification); and \textbf{(Bottom Right) Preference-Driven}, where targets are defined by user portraits and long-term interests (e.g., focusing on a keyboard for technical writers versus a mouse for gamers, or sweet snacks for teenagers versus coffee for professionals).}
  \label{fig:benchmark}
\end{figure*}

\begin{figure*}[h]
  \centering
  \includegraphics[width=1\linewidth]{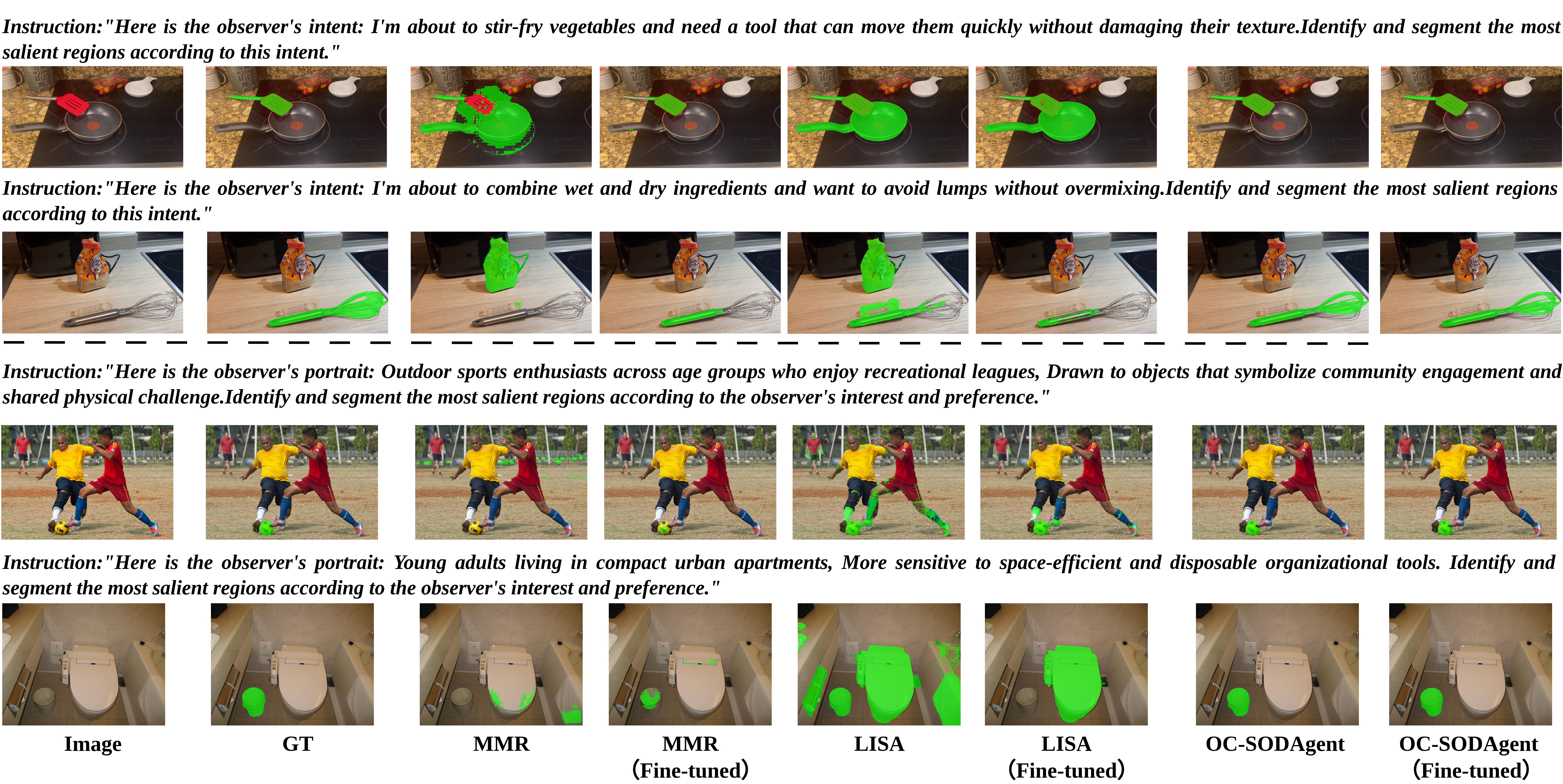}
  \caption{Extended qualitative comparison of visualization results under the intent-driven mode (the top two rows) and preference-driven mode (the last two rows).}
  \label{fig:visualization}
\end{figure*}

\begin{table*}[t]
\centering
\footnotesize  
\setlength{\tabcolsep}{4pt} 
\caption{
    \textbf{Extended Comparison with Fine-Tuned Models.} 
    Evaluation of LISA, MMR, and OC-SODagent on \textbf{Intent-Driven} and \textbf{Preference-Driven} settings. 
    The marked performance leaps of fine-tuned baselines (LISA$_{\text{FT}}$, MMR$_{\text{FT}}$) strongly validate the efficacy of \textbf{OC-SODBench}, while \textbf{OC-SODAgent} maintains a clear lead over these improved baselines, attesting to the robustness of our agent-based design.
}
\label{tab:complexity_guided}
\begin{tabular}{l | ccccc | ccccc}
\toprule
\multirow{2}{*}{\textbf{Model}} & \multicolumn{5}{c|}{\textbf{Intent-Driven}} & \multicolumn{5}{c}{\textbf{Preference-Driven}} \\
\cline{2-11} 
& gIoU  & cIoU  & S$_{m}$  & F$_{m}$  & E$_{m}$  & gIoU  & cIoU  & S$_{m}$  & F$_{m}$  & E$_{m}$  \\
\midrule
\multicolumn{11}{l}{\textit{MLLMs in training-free setting}} \\
\ LISA-7B & 14.53 & 14.83 & 33.68 & 10.86 & 30.90 & 33.02 & 25.49 & 52.28 & 37.66 & 54.73 \\
\ MMR-7B & 11.58 & 11.28 & 47.03 & 14.58 & 56.22 & 19.08 & 15.54 & 46.24 & 24.87 & 57.70 \\
\textbf{OC-SODagent} & 26.23 & 28.83 & 62.90 & 39.97 & 69.23 & 60.61 & 60.54 & 81.08 & 73.97 & 85.79 \\
\midrule
\multicolumn{11}{l}{\textit{MLLMs with finetuning}} \\
\ LISA-7B$_{\text{FT}}$ & 27.18 & \textbf{33.09} & 56.60 & 25.92 & 63.52 & 59.66 & 42.90 & 75.19 & 66.72 & 78.97 \\
\ MMR-7B$_{\text{FT}}$ & 20.50 & 26.79 & 56.05 & 27.54 & 65.77 & 55.44 & 55.75 & 75.29 & 63.39 & 75.96 \\
\textbf{OC-SODagent$_{\text{FT}}$} & \textbf{29.86} & 32.17 & \textbf{65.28} & \textbf{41.21} & \textbf{72.56} & \textbf{62.82} & \textbf{63.99} & \textbf{83.10} & \textbf{75.53} & \textbf{87.69} \\
\bottomrule
\end{tabular}
\end{table*}

\section*{Appendix B. OC-SODAgent Input Prompts}
\addcontentsline{toc}{section}{Appendix B. OC-SODAgent Input Prompts}
\label{app:B}


\subsection*{B.1 OC-SODAgent Input Prompt Templates} \label{app:B1}

\noindent\textbf{Description.} This subsection presents the prompt templates used to drive the \textbf{OC-SODAgent} (refer to Sec.~4, ``Agentic Baseline for OC-SOD'', in the main text). The agentic workflow consists of two distinct phases: (1) an Initial Prediction Stage for generating bounding boxes and referring expressions, and (2) a Perceive--Reflect--Adjust Stage for self-correction.

\paragraph{Stage 1: Initial Prediction} The prompt used in this stage adapts dynamically to the specific observation mode defined in the benchmark (\cref{fig:agent_stage1}): \begin{itemize} \item \textbf{Free-Viewing Mode:} The agent utilizes a general \textbf{Salient Object Localization} prompt. It is instructed to identify objects based on objective visual prominence (e.g., contrast, texture) without specific task constraints. \item \textbf{Intent-Driven \& Preference-Driven Modes:} The agent employs an \textbf{Observer-Centric} localization prompt. Here, the identification is strictly conditioned on the subjective \textit{intent} or \textit{preference} description provided in the instruction, requiring the model to locate targets that align with the specific semantic needs of the observer. \end{itemize}

\paragraph{Stage 2: Perceive--Reflect--Adjust} As illustrated in \cref{fig:agent_stage2}, this stage employs a universal self-correction prompt applicable to all modes. The prompt defines clear \textbf{Visual Input Definitions}: \begin{itemize} \item \textbf{Green Bounding Boxes:} The model's prediction from the previous round. \item \textbf{Red Borders:} The segmentation result generated by the external tool (SAM 2) based on the green boxes. \end{itemize} The agent is instructed to evaluate the alignment between the red segmentation borders and the target object (whether defined by saliency or intent). If the segmentation fails criteria such as \textit{Container Exclusion}, \textit{Occlusion Awareness}, or \textit{Semantic Alignment}, the agent outputs corrected bounding boxes and a reflection analysis.


\section*{Appendix C. OD-SODBench Dataset Demonstration}
\addcontentsline{toc}{section}{Appendix C. OD-SODBench Dataset Demonstration}
\label{app:C}
Representative examples from the OD-SODBench dataset illustrating the three distinct Observer-Centric modes: Free-Viewing, Intent-Driven, and Preference-Driven. As shown in \cref{fig:benchmark}, the dataset challenges models to adapt focus targets based on varying Observer-Centric instruction types for the same visual input:
\begin{itemize}
    \item \textbf{Free-Viewing:} Instructions focus on inherent visual saliency, requiring the model to identify regions based on visual context, color contrast, and semantic meaning (e.g., identifying a prominent animal or vehicle).
    \item \textbf{Intent-Driven:} Segmentation is guided by immediate, task-specific user goals. For instance, distinguishing between a guitar for ``creative engagement" versus a printer for ``document verification" within the same workspace.
    \item \textbf{Preference-Driven:} Targets are defined by user portraits and long-term interests. Examples include differentiating salient objects for a ``competitive gamer" (mouse focus) versus a ``technical writer" (keyboard focus), or shifting attention between food items based on demographic preferences (e.g., sweet snacks for teenagers vs. coffee for professionals).
\end{itemize}

\section*{Appendix D. Additional Experimental Results}
\addcontentsline{toc}{section}{Appendix D. Additional Experimental Results}
\label{app:D}
\subsection*{D.1 Results of Additional Fine-Tuned Models}
We extend our quantitative analysis by including the MMR model to offer a broader perspective on model capabilities. Table~\ref{tab:complexity_guided} presents the performance of LISA, MMR, and OC-SODAgent under both training-free and fine-tuned settings. 
The results highlight the dual effectiveness of our contributions:
(1) The marked performance leaps of LISA-7B$_{\text{FT}}$ and MMR-7B$_{\text{FT}}$ after fine-tuning strongly validate the quality and efficacy of the \textbf{OC-SODBench} dataset in unlocking model potential.
(2) Meanwhile, \textbf{OC-SODAgent} maintains a clear lead over these improved baselines. Its superior performance, particularly in the fine-tuned setting, attests to the robustness of our agent-based design, proving it to be a more optimal solution for complex intent and preference understanding than standard MLLM adaptations.


\subsection*{D.2 Additional Visualization and Comparative Analysis}

We provide extended qualitative examples to substantiate the results in Fig.~5 and Section~5.3. As shown in \cref{fig:visualization}, the visualization is categorized into two distinct scenarios: Intent-Driven (Rows 1-2) and Preference-Driven (Rows 3-4).

To demonstrate the validity and effectiveness of the OC-SODBench dataset, we explicitly compare baseline models (MMR, LISA) and our proposed method (OC-SODAgent) across two states: pre-trained (Vanilla) and fine-tuned (FT) on our dataset. As observed in the visual results, the fine-tuned variants (MMR$_{\text{FT}}$, LISA$_{\text{FT}}$, and OC-SODAgent$_{\text{FT}}$) show marked improvements in locating relevant objects compared to their vanilla counterparts, validating the necessity and quality of the OC-SODBench data.

Notably, even when compared against these fine-tuned baselines, \textbf{OC-SODAgent$_{\text{FT}}$} demonstrates superior performance. It exhibits more precise segmentation and robust reasoning capabilities, accurately identifying the most salient regions aligned with complex observer intents (e.g., selecting the correct spatula for ``stir-frying'' vs. ``mixing'') and personas, whereas other models often suffer from over-segmentation or semantic drift.